\documentclass[journal,comsoc]{IEEEtran}
\usepackage[T1]{fontenc}
\usepackage{amsmath}
\usepackage[cmintegrals]{newtxmath}
\usepackage{bm}
\usepackage{graphicx}
\usepackage{multirow}
\usepackage{floatrow}
\usepackage{comment}
\usepackage{hyperref}
\usepackage{color}

\usepackage{bm}
\usepackage{mathtools}
\usepackage{booktabs}
\usepackage{threeparttable}
\usepackage{tabularx}
\newcolumntype{Y}{>{\centering\arraybackslash}X}
\begin{document}
\title{Towards Reliable Respiratory Disease Diagnosis Based on Cough Sounds and Vision Transformers}
\author{Qian~Wang, 
        Zhaoyang~Bu,
        Jiaxuan~Mao,
        Wenyu~Zhu, 
        Jingya~Zhao,
        Wei~Du,
        Guochao~Shi,
        Min~Zhou,\\
        Si~Chen*,
        Jieming~Qu*
\thanks{Q. Wang, Z. Bu, J. Mao, W. Zhu and S. Chen are with Luca Healthcare, Shanghai, China. (E-mail: qian.wang173@hotmail.com, \{buzhaoyang, mary.mao, wendy.zhu, echo.chen\}@lucahealthcare.com)}
\thanks{J. Zhao, W. Du, G. Shi, M. Zhou and J. Qu are with the Department of Pulmonary and Critical Care Medicine, Ruijin Hospital, Shanghai Jiao Tong University School of Medicine, Shanghai, 200025, China; Institute of Respiratory Disease, Shanghai Jiaotong University School of Medicine, Shanghai, China; Shanghai Key Laboratory of Emergency Prevention, Diagnosis and Treatment of Respiratory Infectious Disease, Shanghai, China. (E-mail: jingya2010@126.com, duweiwilson@qq.com, shiguochao@hotmail.com, doctor\_zhou\_99@163.com, jmqu0906@163.com)}
\thanks{* Corresponding authors.}
}

\markboth{IEEE Trans 2024}%
{Wang \MakeLowercase{\textit{Wang et al.}}: Towards Reliable Respiratory Disease Diagnosis Based on Cough Sounds and Vision Transformers}
\maketitle

\begin{abstract}
Recent advancements in deep learning techniques have sparked performance boosts in various real-world applications including disease diagnosis based on multi-modal medical data. Cough sound data-based respiratory disease (e.g., COVID-19 and Chronic Obstructive Pulmonary Disease) diagnosis has also attracted much attention. However, existing works usually utilise traditional machine learning or deep models of moderate scales. On the other hand, the developed approaches are trained and evaluated on small-scale data due to the difficulty of curating and annotating clinical data on scale. To address these issues in prior works, we create a unified framework to evaluate various deep models from lightweight Convolutional Neural Networks (e.g., ResNet18) to modern vision transformers and compare their performance in respiratory disease classification. Based on the observations from such an extensive empirical study, we propose a novel algorithm \textit{Cough Search} for cough-based disease classification based on both self-supervised and supervised learning on a large-scale cough data set. Experimental results demonstrate our proposed approach outperforms prior arts consistently on two benchmark datasets for COVID-19 diagnosis and a proprietary dataset for COPD/non-COPD classification with an AUROC of 92.5\%. 
\end{abstract}


\IEEEpeerreviewmaketitle

\section{Introduction}
\label{sec:intro}
\IEEEPARstart{R}{espiratory} diseases such as COVID-19 and COPD are widespread in populations and early diagnosis is imperative for disease control and human well-being. According to \cite{mettananda2024burden}, COPD was the third fatal disease in the world in 2022 and is expected to remain the third
cause of death through 2050 at the global level. The current diagnosis of COPD requires measurements of lung functions using clinical equipment in hospitals which is expensive and inconvenient for mass screening. As a result, a significant fraction of COPD patients are undiagnosed in China and are exposed to risks of developing into fatal cancers.

The advancement of machine learning and deep learning techniques enables digital diagnosis for various diseases with satisfying accuracy \cite{piccialli2021survey}. These data-driven techniques can discover hidden patterns in patient's medical data and connect them with disease labels of interest. In general, machines are expected to understand medical data at an equivalent level as experts for disease diagnosis after being exposed to large-scale high-quality data for training.
Nowadays, various medical data modalities have been investigated for disease diagnosis. Comprehensive surveys have been made for deep learning techniques used in medical applications using medical imagery \cite{zhou2023deep}, physiological data \cite{rim2020deep}, electronic health records \cite{solares2020deep} and audio data \cite{sfayyih2023review}.

Among many respiratory sounds such as wheezing, crackles, and breathing, cough sounds have been used to diagnose respiratory diseases in a long history as cough sounds contain indicative patterns closely related to pathomorphological alterations in the respiratory system. Researchers have tried to develop various digital methods for cough sound-based respiratory disease diagnosis \cite{sharan2023detecting,ijaz2022towards}. However, most existing studies use traditional machine learning methods and handcrafted features extracted by classic audio signal processing techniques or relatively lightweight deep models. This is mainly caused by the lack of large-scale cough data sets to pre-train or even fine-tune modern deep models like wav2vec \cite{baevski2023efficient} and transformers \cite{lin2022survey}. For the same reason, the performance reported on small datasets may be biased and unreliable.

Recently, Zhang et al. \cite{zhang2024openrespiratoryacousticfoundation} released a series of respiratory acoustic foundation models which are based on the latest self-supervised learning and are pre-trained on large-scale unlabeled respiratory sound data. In addition, the pre-trained foundation models are evaluated on a curated list of downstream tasks for respiratory sound classification or regression. The benchmarking in this work compares several modern deep models originally designed for audio event classification and demonstrates the superiority of their proposed models which are pre-trained specifically on respiratory sound data. However, their released models are based on backbone models of moderate sizes compared with the most advanced ones for audio event or image classification.

In this work, we focus on cough sound data and investigate the best practices for improving the accuracy of cough-based respiratory disease diagnosis. Motivated by existing works in general audio and image classification, we raise the following research questions in this study: 
 {(1) Which pre-trained models are most appropriate for cough-based respiratory disease classification in terms of accuracy and efficiency?}
 {(2) What are the best practices for fine-tuning a pre-trained model for enhanced cough-based respiratory disease classification?}
 {(3) What are the best practices of model ensemble and selection for optimal classification performance?}

\begin{figure*}
    \centering
    \includegraphics[width=\linewidth]{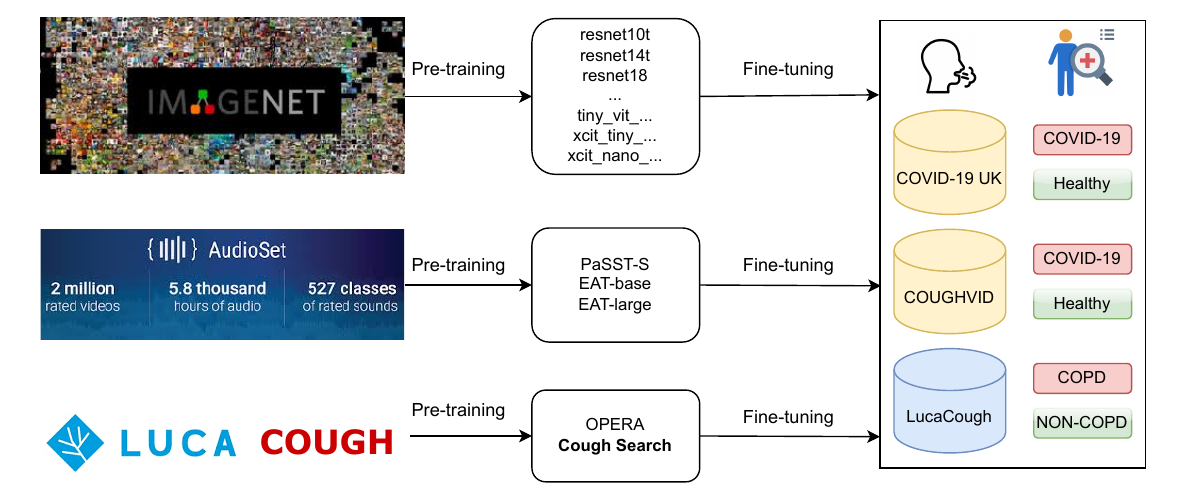}
    \caption{The overview pre-training and fine-tuning pipelines using various datasets, backbone models and downstream tasks in this study.}
    \label{fig:overview}
\end{figure*}

We conduct extensive empirical studies to answer these research questions and propose a novel cough-based respiratory disease classification approach \textit{Cough Search}. The contributions of our work can be summarised as follows:
\begin{itemize}
\item We create a unified framework to investigate the cough-based disease classification performance of various pre-trained models belonging to three categories: ImageNet-pre-trained models, audio data pre-trained models and respiratory sound data pre-trained models.
\item We propose an approach to cough-based respiratory disease classification based on both self-supervised learning and supervised learning on a large-scale cough data set.
\item We conduct extensive experiments on three datasets (two public datasets for COVID-19 classification and one proprietary dataset for COPD/non-COPD classification) and the experimental results demonstrate our proposed approach outperforms all others consistently on three tasks.
\end{itemize}

\section{Related Work} \label{sec:related}
In this section, we review existing works closely related to this study. Specifically, we first review the latest deep learning approaches to audio event classification as our employed method is based on and adapted from models for general audio event classification tasks. Subsequently, we review recent works on cough sound classification and discuss the limitations of commonly used approaches to cough sound classification for respiratory disease diagnosis.

\subsection{Audio event classification} \label{sec:audio_cls}
Audio event classification is a well-formulated research task that has attracted significant attention in the community. Deep learning models have dominated state-of-the-art approaches to this task in recent years. These approaches follow a similar framework in which the raw audio data are converted to log-mel-spectrogram images. Hence, innovations in image classification models can also benefit the audio event classification tasks by proper transfer learning. Such transfer learning is enabled by fine-tuning the pre-trained image classification models on large-scale audio datasets like AudioSet \cite{gemmeke2017audio}. 

According to the model architectures, existing approaches to audio event classification in the literature can be categorized into two groups: Convolutional Neural Networks (CNN) \cite{palanisamy2020rethinking,kong2020panns,drossos2020sound,abrol2020learning} and transformers \cite{gong2021ast}. Whilst early works employing ImageNet-pretrained CNN models on audio spectrograms achieved competitive accuracy, they have been outperformed by transformer-based models. Audio Spectrogram Transformer (AST) \cite{gong2021ast} was one of the first to apply vision transformers to audio event classification and performed superior to their CNN counterparts. Koutini et al. \cite{koutini2021efficient} proposed the PaSST series by evaluating different variants of tailored ViT models and efficient training strategies for audio event classification. These works vary from one another in their employed backbone architectures and audio data preprocessing but share a similar transfer learning framework in which the ImageNet-pre-trained models are fine-tuned on audio spectrogram data for downstream tasks.

With access to large-scale audio datasets like AudioSet-2M, more recent works turn to self-supervised learning for enhanced audio representation learning. A framework consisting of self-supervised learning and supervised fine-tuning has been employed by \cite{li2024self,chen2024eat,alex2024dtf} and leads to state-of-the-art methods for audio event classification in various downstream tasks. 

The aforementioned models are usually evaluated on several benchmark datasets for audio event classification. Those performing well in general audio classification tasks are expected to perform well in a specific downstream task like cough sound classification. We believe a thorough evaluation is imperative for choosing the most appropriate solutions to the specific cough sound classification task in practice. More recently, Zhang et al. \cite{zhang2024openrespiratoryacousticfoundation} made a thorough evaluation of self-supervised learning models across several respiratory sound classification tasks, however, their work aims at a wider scope than cough data classification and their released models were only pre-trained on respiratory sound data which lead to sub-optimal performance as observed in our comparative study.
%

\subsection{Cough sound classification}
Cough sound-based respiratory disease classification has attracted much attention. Researchers have explored the possibility and efficacy of using cough data to diagnose various respiratory diseases including tuberculosis (TB) \cite{pahar2021automatic,sharma2024tbscreen}, Asthma \cite{barata2020automatic,xu2021listen2cough,barua2024automated}, pneumonic infections \cite{kumar2021deep}, and COVID-19 \cite{vrindavanam2021machine,melek2022identifying,xue2021exploring,manzella2023voice}.

However, most of these prior works use traditional machine learning approaches such as logistic regression \cite{vrindavanam2021machine}, support vector machine \cite{melek2022identifying,vrindavanam2021machine}, tree-based models \cite{vrindavanam2021machine}, multi-layer perceptrons \cite{pahar2021covid} and classical convolutional neural networks like ResNet50 \cite{pahar2021covid}. The evaluations were usually made on small private datasets, preventing a direct fair comparison across different works and making the practical use of proposed methods difficult. 
For example, Pahar et al. \cite{pahar2021automatic} evaluated various traditional machine learning methods for cough-based TB classification and the best logistic regression achieves an area under the receiver operator curve (AUROC) of 0.94 using 23 features selected from a set of 78 high-resolution mel-frequency cepstral coefficients. Later, the same group employed the Resnet50 classifier, and discriminated between the COVID-19 positive and the healthy coughs with an area under the ROC curve (AUC) of 0.98 \cite{pahar2021covid}. 

Along with the release of large-scale cough data sets including COUGHVID \cite{orlandic2021coughvid} which contains over 25,000 crowdsourced cough recordings and Coswara \cite{sharma2020coswara} which contains more than 7,000 audio samples from around 1,000 participants for COVID-19 diagnosis. Attempts have also been made using modern deep-learning models for cough classification. Approaches falling into this category preprocess the raw audio data into log-mel spectrogram image data so that they can be fed into deep models originally designed for image classification. Xue et al. \cite{xue2021exploring} propose a novel self-supervised learning framework for COVID-19 cough classification. A vision transformer (ViT) is firstly trained on unlabeled cough data in a self-supervised learning manner and the pre-trained model is subsequently fine-tuned on the downstream classification task for COVID-19 screening.  Valdes et al. \cite{valdes2022cough} also employ a ViT-based model Audio Spectrogram Transformer (AST) \cite{gong2021ast} for cough signal feature extraction towards the classification cough types (e.g., dry, wet, whooping, etc.). The employed AST was pre-trained on a large-scale image dataset ImageNet \cite{deng2009imagenet} and a large-scale audio dataset AudioSet \cite{gemmeke2017audio} subsequently. However, features extracted from the pre-trained models are directly used in the downstream task. We believe the data distribution gap between general audio data (e.g., those in AudioSet) and cough data will restrict the capabilities of deep models without proper transfer learning. To address this limitation, in this study, we further fine-tune the ViT-based deep models on cough data to enhance their representation learning from cough data to discover disease signatures underlying cough sound data.

Although most existing works use the off-the-shelf deep models, Dentamaro et al. \cite{dentamaro2022auco} proposed a novel deep model, dubbed AUCO ResNet, by designing a trainable Mel-like spectrogram layer able to finetune the Mel-like-Spectrogram for capturing relevant time-frequency information. However, it is unclear if such a layer can still benefit ViT-based models when they are pre-trained and fine-tuned on large-scale audio datasets (e.g., AudioSet).

Distinct from prior works on cough-based respiratory disease classification, in this work, we follow \cite{zhang2024openrespiratoryacousticfoundation} and use modern deep learning techniques adapted from the state-of-the-art deep models in audio event classification (c.f. \ref{sec:audio_cls}). The conclusions drawn from our experiments are expected to provide insights into the practical use of cough data for respiratory disease diagnosis.

\section{Method} \label{sec:method}
\begin{figure*}
    \centering
    \includegraphics[width=\linewidth]{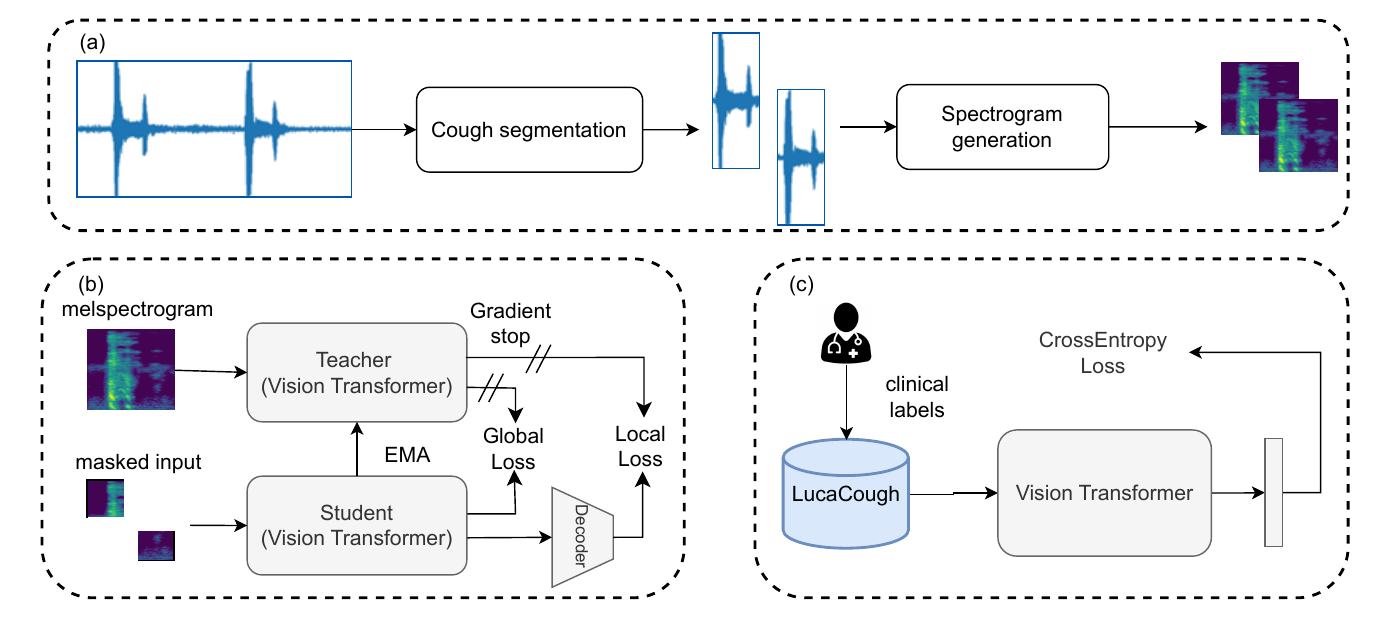}
    \caption{Illustrations of the proposed approach for cough sound-based respiratory disease classification: (a) cough data preprocessing, (b) self-supervised training and (c) supervised fine-tuning on downstream tasks.}
    \label{fig:framework}
\end{figure*}

In this section, we describe the process involved in developing a deep model for cough sound classification. Three steps are illustrated in Figure \ref{fig:framework}: audio data preprocessing (Figure \ref{fig:framework}a), self-supervised learning (Figure \ref{fig:framework}b) and supervised fine-tuning (Figure \ref{fig:framework}c). Whilst the audio data preprocessing step and the supervised fine-tuning step are required by all methods investigated in this study, the self-supervised learning step is only involved in the methods belonging to the respiratory sound data-based pre-training approaches.

\subsection{Cough sound segmentation}
Collecting multiple cough sounds from one subject is practically useful for robust classification results. For this reason, a cough sound segmentation algorithm is needed to segment individual cough sounds from an audio recording. We propose a cough segmentation algorithm based on signal processing and hand-crafted rules.

A cough typically consists of two or three phases \cite{thorpe2001acoustic}. In the first phase, a rapid burst of air breaks through the glottis. In the second phase, the glottis is fully open, allowing air to be steadily expelled from the lungs. Finally, there is possibly a phase where the airflow decreases as the glottis gradually closes.

Our pilot experimental results demonstrate that precise cough onset and offset localisation are not necessary for the classification task. Therefore the cough sound segmentation algorithm boils down to the cough onset detection. A cough event is segmented starting from its detected onset and continues until the segment reaches a predetermined duration.

\textit{Cough onset detection}:
Initially, the audio signal undergoes a Short-Time Fourier Transform (STFT) with a hop length of 0.016 seconds and a window length of 0.021 seconds, utilizing the Hanning window function. Subsequently, the magnitude spectrum is derived by taking the absolute value of the STFT coefficients. Next, the magnitude spectrum is integrated across the 120Hz-8000Hz frequency band to yield a univariate energy sequence. The ratio of the current frame’s energy to that of the preceding frame is computed, and its logarithm is taken to generate a sequence representing the rate of energy change. This sequence is then smoothed using a Butterworth low-pass filter. Peaks in the energy change rate are identified by applying a threshold of 100 to this sequence. For each detected peak in energy change rate, the corresponding energy peak is located within a window of 5 frames centred on the peak. Ultimately, the cough onset is identified as occurring two frames before the located energy peak positions.

\subsection{Spectrogram generation}
It is \textit{de facto} standard to convert raw audio data into 2D spectrogram images for audio event classification using image classification deep models. The generation of spectrogram images is based on STFT and the choice of optimal hyper-parameters is coupled with the type of deep models employed for the classification.

\subsection{Pre-trained models}
In the deep learning era, a common practice is to use a pre-trained deep model and fine-tune it on the downstream task. It is crucial to choose an appropriate pre-trained model for good performance in the downstream tasks. For cough sound spectrogram-based disease classification problems, the candidate pre-trained models can be categorised into three groups according to the pre-training data modalities. In this study, three types of pre-trained models are considered: ImageNet-pre-trained models, audio data pre-trained models and respiratory sound data pre-trained models

ImageNet-pre-trained models have been widely used in a variety of computer vision tasks due to the fact they are extensively studied, rapidly evolved and easily accessible. These models are pre-trained on ImageNet and/or other large-scale natural image datasets of this kind. Prior works have proved the benefit of using such pre-trained models on audio classification tasks \cite{pahar2021covid} provided the models are properly fine-tuned.
As ImageNet-pre-trained models require 3-channel images as inputs, we duplicate the generated 1-channel spectrogram three times to form 3-channel images. For models requiring specified input image sizes (e.g., $224 \times 224$), we resize the images as per requirement.

Audio data pre-trained models close the gap between natural images and audio spectrogram images by pre-training deep neural networks on large-scale audio data. During pre-training, the models take audio spectrogram images as the input and hence can be directly applied to downstream tasks for respiratory sound classification. As the pre-trained models may have employed different parameters to generate the spectrogram, the same process for spectrogram generation as that used during pre-training must be employed during fine-tuning.

Respiratory sound data pre-trained models further close the gap between the general audio spectrogram and the respiratory sound spectrogram. Similar to the audio data pre-trained models, these models can be directly fine-tuned on downstream tasks once the classification head is adapted to fit the number of classes.

A comparative study will be presented in the section on experiments with typical pre-trained models selected from three categories for empirical evaluations.

\subsection{Model fine-tuning on cough data}
To adapt the pre-trained foundation models to a specific downstream task (i.e. cough sound classification in our case) which usually has relatively less labelled data for training, we employ different training paradigms: supervised training and self-supervised training. One individual or a hybrid training strategy can be chosen for the best performance based on the type of training data available in a specific application scenario.

\subsubsection{Supervised training}
In most cases, a supervised training strategy should be used to fine-tune the pre-trained models and adapt them to the downstream task. For this purpose, labelled data are required and the label space is usually different from that in the pre-training phase. As a result, one needs to replace the classification head in the model for fine-tuning. 

The classification heads in modern deep learning architectures are usually implemented with a series of fully connected layers together with non-linear activation and normalisation layers. Whilst the classification heads can vary in the number of layers and the dimension of the input layer (determined by the dimension of the features from the backbone of the pre-trained model), the last layer should always consist of output neurons equalling the number of labels in specific downstream tasks. 


\textbf{SAM}:  Sharpness-Aware Minimization (SAM) \cite{foretsharpness} has been proven as an effective strategy for improving the model generalisation and training efficiency. It seeks parameters that lie in neighbourhoods having uniformly low loss; this formulation results in a min-max optimization problem on which gradient descent can be performed efficiently. Used together with optimizers such as Adam \cite{kingma2014adam} during fine-tuning, it is expected to improve the performance of pre-trained models on the downstream tasks.

\textbf{Data augmentation}: Data augmentation is a commonly used strategy during supervised training to improve the model generalisation and performance. A two-stage data augmentation strategy is employed on the fly during supervised training. In the first stage, we use the Python library \textit{audiomentations} \footnote{https://github.com/iver56/audiomentations} and apply three types of data augmentation to the raw audio data: adding Gaussian noises, multiplying the audio by a random gain factor and shifting the pitch up or down without changing the tempo.
In the second stage, the data augmentation is applied to the spectrogram images. We use \textit{specAugment} \cite{park2019specaugment} and \textit{mixup} \cite{zhang2018mixup} methods to augment the training data in each mini-batch on the fly during training. Specifically, three types of augmentation are applied to the log mel spectrogram images: time warping, frequency masking and time masking.

\textbf{Data imbalance}
It is common to have unbalanced data in medical applications where positive samples (with diseases) are far less than negative ones (healthy). To combat such an issue, we use a simple yet effective strategy which assigns label-dependent weights when computing the loss. The loss for samples belonging to the lower-represented class (positive) is multiplied by a higher weight. The weight can be calculated as the ratio between negative and positive samples in the training set.

\subsubsection{Self-supervised training}
Adapting pre-trained foundation models to a specific downstream task suffers from out-of-distribution issues when the supervised training data are limited. In such cases, self-supervised training on large-scale unlabeled data before the supervised training can bridge the data distribution gap between those used in pre-training and fine-tuning. In our targeted use cases, the models pre-trained either on ImageNet or AudioSet may suffer the out-of-distribution issue when applied to the cough data. To mitigate the performance degradation caused by the distribution gap, we use a two-stage fine-tuning pipeline (c.f. Fig. \ref{fig:framework}) consisting of a self-supervised training stage followed by supervised training on the downstream tasks.

We follow \cite{chen2024eat} and employ the teacher-student framework (c.f. Fig. \ref{fig:framework}b) for self-supervised training on large-scale unlabeled cough data. The teacher and the student models share the same architecture of a vision transformer. The student model weights are updated normally by gradient descent whilst the teach model weights are updated by the Exponential Moving Average (EMA) strategy. The objective of training the student model is composed of a global loss and a local loss. Both are implemented by the Mean Squared Error (MSE) loss. The global loss aims to align the final representations output by the student and teacher models.
\begin{equation}
    \label{eq:lossl}
    L_{local} = ||\bm{X}_s-\bm{F}_t||_2^2
\end{equation}
\begin{equation}
    \label{eq:lossg}
    L_{global} = ||\bm{c}_t-\bm{f}_s||_2^2
\end{equation}

where $\bm{F}_t\in \mathbb{R}^{P'\times M \times E}$ are the local patch representations output by all the transformer blocks in the teacher model and $\bm{X}_s\in \mathbb{R}^{P'\times M \times E}$ is the decoder output which aims to approximate the local representations of the masked patches based on the representations of unmasked patches from the student model; $P'$ is the number masked patches in the input spectrogram to the student model and $M$ is the number of transformer blocks; $\bm{c}_s \in \mathbb{R}^{1\times E}$ is the output representation from the student model and $\bm{f}_t \in \mathbb{R}^{1\times E}$ is the average pooling result of $\bm{Y}_t$; $E$ is the dimension of the representations.  The final loss is a combination of global and local losses.

\section{Experiments and Results} \label{sec:exp}
In this section, we conduct thorough experiments to compare several deep models and experimental settings on classifying respiratory diseases based on cough sounds. We demonstrate the effectiveness of modern deep models in such tasks in practical use cases and provide insights into the choice of best practices in different stages of model training and inference.

\begin{table*}[t]
  \centering
  \caption{Evaluation results (AUROC) of ImageNet pre-trained models}\label{tab:in}
  \small 
  \begin{tabularx}{0.6\linewidth}{@{}l|ccc@{}} 
  \toprule
  & \multicolumn{3}{c}{Dataset} \\ \midrule
  Method & LucaCough & {UK COVID-19} & COUGHVID  \\ \midrule
  resnet10t.c3\_in1k & 84.0 & 64.2 & 56.5  \\
  resnet14t.c3\_in1k & 83.9 & 58.7 & 57.8 \\
  resnet18.a3\_in1k & 76.8 & \it 65.6 & \bf 58.1 \\
  resnet34.a3\_in1k & 71.2 & 62.2 & 52.8 \\
  resnet50.a3\_in1k & 65.8 & 60.5 & 52.8\\
  resnet101.a3\_in1k & 66.4 & 62.9 & 53.8 \\
  \midrule
  tiny\_vit\_5m\_224 & 84.5 & 62.4 & 52.8 \\
  tiny\_vit\_21m\_224 & \bf 85.3 & \bf 66.9 & 57.1  \\
  xcit\_nano\_12\_p8\_224.fb\_dist\_in1k & \it 85.2 & 61.7 & 55.3 \\
  xcit\_tiny\_12\_p8\_224.fb\_in1k & 83.4 & 62.6 & 53.2 \\
  \bottomrule
  \end{tabularx}
\end{table*}

\subsection{Dataset} \label{sec:dataset} 
Three datasets are employed in the experiments for a thorough evaluation across different tasks and data distribution. 

The \textbf{UK COVID-19} dataset \cite{coppock2024audio} is collected to evaluate machine learning models classifying COVID-19 status using multi-modal data including cough sounds. In this study, we use the ``matched" training-test split released along with the dataset \footnote{https://zenodo.org/records/11167750}. As a result, there are 2,599 COVID+ and 2,599 COVID- participants in the training set and 907 COVID+ and 907 COVID- participants in the test set \cite{coppock2024audio}. We follow the same data split of task 2 in \cite{zhang2024openrespiratoryacousticfoundation} and formulate a binary classification task classifying the status of COVID-19  on this dataset.

The \textbf{CoughVID} dataset \cite{orlandic2021coughvid} contains over 25,000 cough sounds labelled with COVID-19 statuses. Following the definition of task 5 in \cite{zhang2024openrespiratoryacousticfoundation}, we use this binary classification task (COVID-19 vs healthy) as one of the test beds in our study.

Our private \textbf{LucaCough} dataset is challenging yet more clinically useful as the negative samples cover not only the coughs of healthy subjects but also the coughs of subjects diagnosed with various respiratory diseases which may share similar cough signatures to COPD cough sounds. Specifically, the negative samples in our dataset include those collected from patients diagnosed with asthma, upper/lower respiratory tract infection (URI/LRI), Bronchiectasis and other non-COPD respiratory diseases. The LucaCough dataset consists of cough data from 3000 subjects (272 COPD and 2728 non-COPD subjects) in the training subset and cough data from 651 subjects (66 COPD and 586 non-COPD) in the test set. The cough sounds are recorded via several different mobile phones to make the data more diverse and enable the generalisation capabilities of the developed models. The data are manually annotated by at least three clinical experts with the help of medical records and lung function test readings. Informed written consent was obtained from all participants. The study was approved by the Ruijin Hospital Shanghai Jiaotong University School of Medicine Ethics Committee (No. 2023-199), Ruijin Hospital Luwan Branch Ethics Committee (2023-HXK-V1), Shanghai Jing'an District Central Hospital Ethics Committee (No. 2023-33), Shanghai Zhabei Central Hospital Ethics Committee (ZBLL2024030401001) and registered at ClinicalTrials.gov (NCT06082791).

\subsection{Implementation details} \label{sec:metric}
We implement the proposed method in PyTorch and all the experiments are run on a GeForce RTX 4090 GPU. In the self-supervised pre-training stage, we use the default experimental settings used in \cite{chen2024eat}. For supervised fine-tuning, we use a learning rate of 2e-6 and a batch size of 24 throughout our study. The Adam optimizer is employed if not specified otherwise.

Since all downstream tasks investigated in our experiments are binary classification problems, we use the area under the receiver operator curve (AUROC) as the evaluation metric if not otherwise specified. Multiple runs with different random seeds are conducted for each experiment to get the mean values as reliable results.

\subsection{Evaluation of ImageNet pre-trained models} \label{sec:arch}
There exist plenty of deep models pre-trained on ImageNet. In this study, we use the \textit{timm} library \footnote{https://github.com/huggingface/pytorch-image-models} as a convenient tool to access various pre-trained vision models in a unified framework. Firstly, we choose the ResNet series from ResNet10 to ResNet101 to investigate how the model complexity of Convolutional Neural Networks (CNN) affects the performance on three downstream tasks. The results are shown in Table \ref{tab:in}. It is interesting to see models with fewer layers (e.g., resnet10t, resnet14t and resnet18) generally outperform those with more layers consistently on three datasets. The possible reason could be deeper models require more training data to achieve good performance than their shallow counterparts.

Based on the observations of ResNet performance, we conduct follow-up experiments on four small vision transformer models as listed in the bottom part of Table \ref{tab:in}. These transformer-based models perform comparably well with their CNN counterparts and the best one achieves the highest AUROC on two tasks (i.e. LucaCough and UK COVID-19), the third highest on the COUGHVID dataset.

Experimental results shown in Table \ref{tab:in} demonstrate deeper neural networks with more model capacities may not always be better choices for downstream tasks. Our empirical study provides insight into how to choose ImageNet-pre-trained models for downstream tasks when the data distribution gap is large (e.g., natural images and cough audio spectrogram) and the amount of data for fine-tuning is limited.

\subsection{Evaluation of AudioSet pre-trained models} \label{sec:ssl}
In this section, we evaluate the performance of models pre-trained on audio data (e.g., AudioSet-2M). Specifically, we evaluate PaSST-S \cite{koutini2021efficient}, EAT-base and EAT-large \cite{chen2024eat} on the benchmark datasets. PaSST-S is based on DeiT-B384 \cite{touvron2021training} which has the same architecture as ViT-B \cite{dosovitskiy2020image}. EAT-base is based on ViT-B hence both PaSST-S and EAT-base have 86M parameters, 12 Multi-head self-attention (MSA) layers, 12 attention heads and the embedding dimension is 768. EAT-large is based on ViT-L and has 307M parameters, 24 MSA layers, 16 attention heads and the embedding dimension is 1024.

We use the code and checkpoints released by the authors of these two works and follow their instructions for fine-tuning on the three downstream tasks. All the checkpoints utilised in our experiments were trained on the large-scale AudioSet-2M dataset in a supervised learning way. In addition, the EAT models were also pre-trained on the AudioSet-2M dataset using a self-supervised learning strategy before fine-tuning on the same dataset. To further fine-tune the models on our downstream tasks, the classification head is replaced with a linear layer with two output neurons for the binary classification problems.

The experimental results are shown in Table \ref{tab:as}. The investigated models perform comparably well on three downstream tasks. On the COUGHVID dataset, EAT-large achieves higher AUROC than the lighter version EAT-base which again outperforms PaSST-S. However, compared with the experimental results shown in Table \ref{tab:in}, all three audio data pre-trained models perform significantly better than most of the ImageNet-pre-trained ones on three downstream tasks consistently. The comparison between experimental results in Tables \ref{tab:in} and \ref{tab:as} provides clear evidence that pre-training on audio data benefits the classification of respiratory data including cough sounds.

\begin{table}[t]
  \centering
  \caption{Evaluation results (AUROC) of models pre-trained one audio data (upper part) and respiratory sound data (bottom part)}\label{tab:as}
  \small 
  \begin{tabularx}{\linewidth}{@{}l|ccc@{}} 
  \toprule
  & \multicolumn{3}{c}{Dataset} \\ \midrule
  Method & LucaCough & {UK COVID-19} & COUGHVID  \\ \midrule
  CLAP \cite{elizalde2023clap} & - & 64.8 & 59.9 \\
  OPERA-CT \cite{zhang2024openrespiratoryacousticfoundation} & - & 70.1 & 57.8  \\
  OPERA-CE \cite{zhang2024openrespiratoryacousticfoundation} & - & 62.9 & 56.6  \\
  OPERA-GT \cite{zhang2024openrespiratoryacousticfoundation} & - & 67.7 & 55.2  \\
  \midrule  
  PaSST-S \cite{koutini2021efficient} & 88.2 & 70.5 & 56.5  \\
  EAT-base \cite{chen2024eat} & 88.2 & 70.3 & 57.7  \\
  EAT-large  \cite{chen2024eat}& 88.4 & 70.4 & 58.9  \\
  \midrule
  OPERA-CT \cite{zhang2024openrespiratoryacousticfoundation} & 85.2 & 69.4 & 58.4  \\
  OPERA-CE \cite{zhang2024openrespiratoryacousticfoundation} & 83.0 & 65.5 & 57.5  \\
  OPERA-GT \cite{zhang2024openrespiratoryacousticfoundation} & 86.7 & 69.2 & 57.7  \\
  \textit{Cough Search} (Ours) & \bf 91.7 & \bf 71.3 & \bf 60.7  \\
  \bottomrule
  \end{tabularx}
\end{table}

\subsection{Evaluation of Respiratory sound data pretrained models} \label{sec:ft}
We conduct experiments to evaluate models pre-trained on respiratory sound data. The models employed in these experiments include those released by Zhang et al. \cite{zhang2024openrespiratoryacousticfoundation} and EAT models further trained on cough data by ourselves. Firstly, we apply the released pre-trained models to the LucaCough datasets and present the experimental results together with those from \cite{zhang2024openrespiratoryacousticfoundation} on the other two datasets. Subsequently, we continue the pertaining of EAT models on a large-scale cough sound dataset. As a result, we have a variant of the pre-trained EAT-large model dubbed \textit{Cough Search} as it was further trained on cough data. The \textit{Cough Search} model is subsequently fine-tuned on downstream tasks to obtain the classification results and compare them with other respiratory sound data pre-trained models (i.e. the OPERA series). To make a fair comparison, we use the official code released by \cite{zhang2024openrespiratoryacousticfoundation} to fine-tune the OPERA models on downstream tasks whilst only linear probing results were reported in the original paper \cite{zhang2024openrespiratoryacousticfoundation}.

The comparison results are shown in Table \ref{tab:as} from which several conclusions can be drawn. Firstly, \textit{Cough Search} performs the best on three tasks consistently. Particularly, it outperforms EAT-large on three tasks with the AUROC margins ranging from 0.9 to 3.3 percentage points. This proves that continuous pre-training of the model on cough data can further strengthen its capabilities of classifying diseases based on cough sound data.  Secondly, the OPERA models, though pre-trained on respiratory sound data in a self-supervised learning manner, perform inferior not only to \textit{Cough Search} but also to the models pre-trained on general audio data (i.e. PaSST-S, EAT-base and EAT-large). This may be attributed to the fact of low model capacities or the lack of pre-training on large-scale audio data like AudioSet-2M.


\subsection{On the effectiveness of SAM} \label{sec:exp_sam}
In this subsection, we focus on exploring the effectiveness of the SAM optimizer in our particular cough classification tasks. Specifically, we compare experimental results without and with the use of SAM optimizer during fine-tuning whilst keeping all other experimental settings the same. The comparative study is conducted on the LucaCough dataset with three representative pre-trained models: PaSST-S, EAT-base, EAT-large and \textit{Cough Search}. The empirical results are shown in Table \ref{tab:sam}. The experimental results demonstrate the use of SAM improves the AUROC values consistently for all four models.

\begin{table}[t]
  \centering
  \caption{Experimental results on LucaCough without and with the use of SAM}\label{tab:sam}
  \begin{tabularx}{0.8\linewidth}{@{}l|cc} 
  \toprule
  Model & w/o SAM & w/ SAM  \\ \midrule
  PaSST-S \cite{koutini2021efficient} & 88.2 & 89.1 $\uparrow$ \\
  EAT-base \cite{chen2024eat} & 88.2 & 89.3 $\uparrow$ \\
  EAT-large \cite{chen2024eat} & 88.4 & 89.8 $\uparrow$  \\
  \textit{Cough Search} (Ours) &  91.7 &  92.5 $\uparrow$  \\
  \bottomrule
  \end{tabularx}
\end{table}


\section{Conclusions and Discussion} \label{sec:conclusion}
This study presents a comprehensive evaluation of various deep learning models and their performance in classifying respiratory diseases such as COVID-19 and Chronic Obstructive Pulmonary Disease (COPD) using cough sound data. Our proposed approach, which leverages both self-supervised and supervised learning on a large-scale cough dataset, has demonstrated superior performance on three datasets.

The use of pre-trained models, particularly those pre-trained on respiratory sound data, has shown significant benefits in enhancing the classification of cough sounds. The continuous pre-training of models on cough data, as evidenced by the performance of the \textit{Cough Search} model, further strengthens the model's ability to discern disease signatures from cough sound data.

While our approach has shown promising results, there is room for further improvement and exploration. Future work could involve the integration of additional data modalities, such as clinical and demographic information, to enhance the predictive power of the models. The development of more sophisticated self-supervised learning techniques tailored to the unique characteristics of cough sounds could potentially uncover more nuanced patterns in the data.

In conclusion, our study represents a significant step towards reliable and accurate respiratory disease diagnosis using cough sounds. The insights gained from this research have the potential to inform the design of future diagnostic systems, ultimately contributing to improved disease control and human well-being.
\bibliographystyle{IEEEtran}
\bibliography{ref}







\end{document}